%% file: main.tex
\documentclass[letterpaper]{article} 
\usepackage{aaai2026}  
\usepackage{times}  
\usepackage{helvet}  
\usepackage{courier}  
\usepackage[hyphens]{url}  
\usepackage{graphicx} 
\usepackage{natbib}  
\usepackage{caption} 
\usepackage{algorithm}
\usepackage{algorithmicx}
\usepackage{algpseudocode}
\usepackage{colortbl}
\usepackage{array}
\usepackage[dvipsnames]{xcolor}
\usepackage{multirow}
\usepackage{amssymb}
\usepackage{makecell}
\usepackage{amsmath}
\usepackage{amsfonts}
\usepackage[listings]{tcolorbox}
\usepackage{booktabs}
\usepackage{subcaption}
\definecolor{mypink}{HTML}{FEB5AF}
\definecolor{mygray}{gray}{.9}
\usepackage{tabularx}
\newcolumntype{Y}{>{\setlength\parindent{2em}}X}

\usepackage{pifont}
\newcommand{\cmark}{\ding{51}}%
\newcommand{\xmark}{\ding{55}}%

\usepackage{newfloat}
\usepackage{listings}
\DeclareCaptionStyle{ruled}{labelfont=normalfont,labelsep=colon,strut=off} 
\floatstyle{ruled}
\newfloat{listing}{tb}{lst}{}
\floatname{listing}{Listing}
\newtheorem{dfn}{Definition}

\title{Bolster Hallucination Detection via Prompt-Guided Data Augmentation}
\author{
    Wenyun Li\textsuperscript{\rm 1,2},
    Zheng Zhang\textsuperscript{\rm 1,2 }\thanks{Corresponding authors.},
    Dongmei Jiang\textsuperscript{\rm 2 },
    Xiangyuan Lan\textsuperscript{\rm 2,3}
}
\affiliations{
    \textsuperscript{\rm 1}Harbin Institute of Technology, Shenzhen, China\\  
    \textsuperscript{\rm 2}Pengcheng Laboratory , Shenzhen, China\\
    \textsuperscript{\rm 3}Pazhou Laboratory (Huangpu), Guangzhou, China\\
    liwy@pcl.ac.cn, darrenzz219@gmail.com, jiangdm@pcl.ac.cn, lanxy@pcl.ac.cn
}

\usepackage{bibentry}

\begin{document}

\maketitle

\begin{abstract}
Large language models (LLMs) have garnered significant interest in AI community. Despite their impressive generation capabilities, they have been found to produce misleading or fabricated information, a phenomenon known as hallucinations. Consequently, hallucination detection has become critical to ensure the reliability of LLM-generated content. One primary challenge in hallucination detection is the scarcity of well-labeled datasets containing both truthful and hallucinated outputs. To address this issue, we introduce \textbf{P}rompt-guided data \textbf{A}ugmented ha\textbf{L}lucination d\textbf{E}tection (PALE), a novel framework that leverages prompt-guided responses from LLMs as data augmentation for hallucination detection. This strategy can generate both truthful and hallucinated data under prompt guidance at a relatively low cost. To more effectively evaluate the truthfulness of the sparse intermediate embeddings produced by LLMs, we introduce an estimation metric called the Contrastive Mahalanobis Score (CM Score). This score is based on modeling the distributions of truthful and hallucinated data in the activation space. CM Score employs a matrix decomposition approach to more accurately capture the underlying structure of these distributions. Importantly, our framework does not require additional human annotations, offering strong generalizability and practicality for real-world applications. Extensive experiments demonstrate that PALE achieves superior hallucination detection performance, outperforming the competitive baseline by a significant margin of 6.55\%.
\end{abstract}

\begin{links}
    \link{Code}{https://github.com/li-wenyun/PALE}
\end{links}

\section{Introduction}
Recently, large language models (LLMs) have emerged as one of the most significant breakthroughs in the field of artificial intelligence \cite{hurst2024gpt}. LLMs have found widespread application across a variety of tasks, including logical reasoning \cite{dong2024insightvexploringlongchainvisual}, visual question answering \cite{jian-etal-2024-large}, and speech-to-text transcription \cite{zhang-etal-2023-speechgpt}, often surpassing human performance in many scenarios. Due to their exceptional reasoning and generative capabilities, LLMs have been integrated into numerous high-trust systems, such as those in the medical domain \cite{liu2023medical}. However, despite these impressive capabilities, even state-of-the-art LLMs frequently generate factually incorrect or nonsensical content—a phenomenon known as hallucination \cite{huang2025survey}. This inherent risk makes LLM outputs potentially unreliable in mission-critical applications \cite{wang2024interactive}. Therefore, a reliable LLM should not only produce text that is coherent with the given prompt but also possess the ability to detect hallucinations in its output.

This concern underscores the importance of hallucination detection technology, which determines whether a generated output is truthful or not \cite{chen2023hallucination,SriramananBSSKF24,su2024unsupervisedrealtimehallucinationdetection}. Most approaches to hallucination detection rely on devising uncertainty scoring functions. For instance, logit-based methods \cite{ren2023outofdistribution,malinin2021uncertainty} employ token-level probabilities as uncertainty scores, consistency-based methods \cite{lin2024generating,manakul2023selfcheckgpt} quantify uncertainty by evaluating the consistency across multiple responses, and verbalized methods \cite{lin2024generating,kadavath2022language} prompt LLMs to express their uncertainty in natural language. Recently, there has been increasing work \cite{burns2022discovering,NEURIPS2024_ba927059,arteaga2024hallucinationdetectionllmsfast} aimed at leveraging the internal states of LLMs to assess the veracity of their outputs. For example, contrast-consistent search (CCS) \cite{burns2022discovering} trains a binary truthfulness classifier to satisfy logical consistency properties, and HaloScope \cite{NEURIPS2024_ba927059} seeks to identify hallucination subspaces. However, all these methods face the challenge of a scarcity of well-labeled datasets containing both truthful and hallucinated outputs, which significantly limits their performance in detecting hallucinations in real-world scenarios.

We argue that the primary challenge in hallucination detection is the lack of labeled datasets containing both truthful and hallucinated generations. In practice, creating a high-quality ground truth dataset for hallucination detection requires substantial human labor to annotate a large number of generated samples. However, collecting such large-scale data and labeling it is extremely costly. In particular, the vast landscape of generative models and the diverse range of content they produce further complicate the process. Moreover, maintaining the quality and consistency of labeled data for hallucination detection requires continuous annotation efforts and robust quality control measures. These formidable obstacles underscore the need to reduce reliance on human-labeled data for hallucination detection.
\begin{figure*}[h]
\centering
\includegraphics[width=1.82\columnwidth]{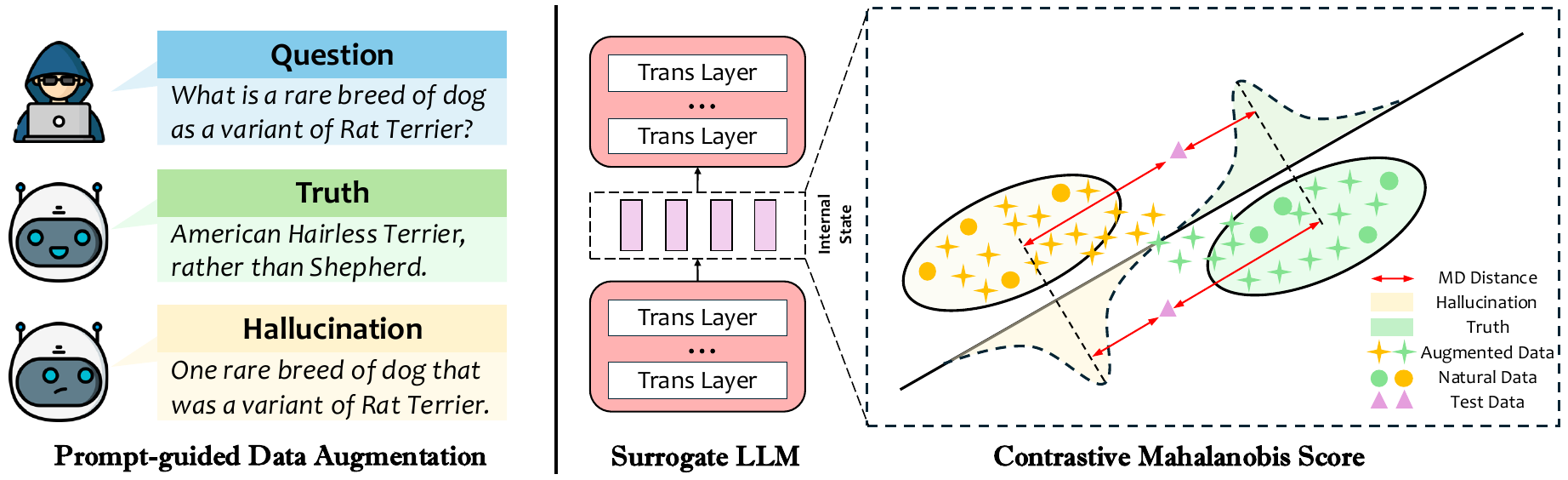}
\vspace{-.7em}
\caption{The pipeline of our proposed PALE. Our PALE comprises three steps: 1)  Utilizing state‑of‑the‑art LLMs to generate both truthful and hallucinated data; 2)  Employing the Contrastive Mahalanobis Score to determine the truthiness of test samples.\label{pipeline}}
\end{figure*}

To address this issue, we introduce \textbf{P}rompt-guided data \textbf{A}ugmented ha\textbf{L}lucination d\textbf{E}tection (PALE), a novel framework that leverages state-of-the-art LLMs for data augmentation using non-parametric prompts. The augmented data generated through LLM prompting can be acquired at a relatively low cost. A state-of-the-art language model, such as GPT-4o, contains a vast amount of hidden knowledge, enabling us to harness LLMs to produce diverse truthful and hallucinated data via prompt engineering. 

From this perspective, we introduce a novel metric, the \textbf{Contrastive Mahalanobis Score}, to assess the truthfulness of augmented data. Our key idea is inspired by prior work \cite{NEURIPS2024_ba927059, burns2022discovering}, which leverages the latent state of a language model as a representation that encodes information related to truthfulness. Specifically, PALE employs a matrix decomposition approach to model the distributions of truthful and hallucinated data in the embedding activation space. It then computes the distance of a test sample to both distributions. Both truthful and hallucinated distributions are modeled as Gaussian, with their means and covariance matrices estimated from the embedding activations. Notably, the covariance matrix computation is facilitated by factorizing the LLM embeddings. The Mahalanobis distance between a test sample and each distribution is used to infer its truthfulness. The resulting score is straightforward to implement in practical applications.

The primary contributions of our research are summarized as follows:
\begin{itemize}
    \item Our proposed framework, \textbf{P}rompt-guided data \textbf{A}ugmented ha\textbf{L}lucination d\textbf{E}tection (PALE), harnesses augmented data generated through prompt engineering for hallucination detection in LLMs. This approach not only yields significant performance improvements but also reduces reliance on costly human annotation.
    \item We present a scoring function called \textbf{Constrastive Mahalanobis Score}, which leverages the distances to both truthful and hallucinated data distributions in the activation space to effectively determine the truthfulness of test data.
    \item Extensive experiments demonstrate the superior effectiveness of our method compared to other state-of-the-art approaches. Moreover, comprehensive ablation studies assess the impact of various design choices in PALE and confirm its scalability across larger LLMs and diverse datasets. These findings provide a systematic and thorough understanding of leveraging LLM-augmented data for hallucination detection, paving the way for future research.
\end{itemize}

\section{Related Works}
\subsection{Hallucination in LLMs}

Large Language Models (LLMs) have made remarkable achievements across various AI domains, including  logical reasoning \cite{dong2024insightvexploringlongchainvisual,wang2024t}, visual question answering \cite{jian-etal-2024-large}, text generation \cite{min2023recent}, and speech-to-text transcription \cite{zhang-etal-2023-speechgpt}. Despite their impressive performance, LLMs still face multiple challenges \cite{lin-etal-2024-towards-understanding}. Notably, one of the most critical failures is the presence of factual errors in generated text, formally referred to as hallucinations \cite{huang2025survey,xu2025hallucinationinevitableinnatelimitation}. The existence of hallucinations poses a significant risk in security-critical scenarios such as medical computer-aided diagnosis \cite{wang2024interactive} and financial decision-making \cite{li2024investorbenchbenchmarkfinancialdecisionmaking}. The issue of hallucinations in natural language generation was recognized by NLP researchers even before the widespread adoption of LLMs \cite{3571730}. First, the large-scale data corpora used to train LLMs inevitably contain erroneous information \cite{shumailov2024ai}, which may contribute to hallucinated outputs. Second, the decoder component of LLMs is typically trained using maximum likelihood estimation, where each token is predicted based on the previously generated sequence, making hallucinations prone to compounding over time \cite{zhang2024how}. Overall, hallucinations in LLMs remain an unresolved issue, calling for further in-depth research.

\subsection{Hallucination Detection}
Hallucination detection \cite{arteaga2024hallucinationdetectionllmsfast,liu2024trustworthyllmssurveyguideline} has recently gained significant attention as a means to address safety concerns in LLMs and ensure their reliability in real-world deployments. A wide range of studies approach hallucination detection by designing uncertainty scoring functions. For instance, logit-based methods \cite{ren2023outofdistribution,malinin2021uncertainty} utilize token-level log probabilities as uncertainty scores, consistency-based methods \cite{lin2024generating,manakul2023selfcheckgpt} assess uncertainty by comparing multiple generated responses, and verbalized methods \cite{lin2024generating,kadavath2022language} prompt LLMs to express confidence in natural language. More recently, internal state-based methods \cite{NEURIPS2024_ba927059} have been proposed, which leverage hidden activations to detect hallucinations. Notable examples include contrast-consistent search (CCS) and HaloScope \cite{NEURIPS2024_ba927059}. However, these methods still suffer from a scarcity of well-labeled datasets containing both truthful and hallucinated generations. For example, CCS relies on manually curated factual datasets, which require substantial human annotation. HaloScope \cite{NEURIPS2024_ba927059}, on the other hand, employs unlabeled data from the same distribution, limiting its generalizability to more diverse and practical scenarios. In contrast, our method performs hallucination detection with minimal human supervision, making it more applicable to real-world settings. It is important to note that our research problem differs from prior work. For data augmentation, we utilize prompt engineering techniques to generate large-scale hallucinated data. The latent knowledge within LLMs can be harnessed to enrich the diversity of hallucination samples.

\section{Methodology}
In this section, we first formally introduce the LLM generation process and define the problem of hallucination detection. We then present the overall framework, followed by a detailed description of the hallucination detector used during inference.
\subsection{Problem Formulation}
\begin{dfn}[\textbf{LLM generation}]
Given an $L$-layer causal LLM, which takes a sequence of tokens $\mathrm{x} =\left \{ x_{1},\dots ,x_{n} \right \}$ as input, an output $\mathrm{y} =\left \{ y_{n+1},\dots ,y_{n+m} \right \} $ with Markov property is generated in an autoregressive manner. Each output token $y_j$, $j \in \left [ n+1, \dots , n+m \right ] $ is sampled from a distribution over the model vocabulary $\mathcal{V} $, conditioned on the prefix $\left \{ x_{1},\dots ,x_{n} \right \}$, which aims maximizing the conditional probability:
\begin{equation}
    y_j= \underset{y \in \mathcal{V} }{\mathrm{argmax}} P\left ( y|\left \{ x_1,\dots,x_{j-1} \right \}  \right ) ,
\end{equation}
and the probability $P$ is calculated as:
\begin{equation}
    P\left ( y|\left \{ x_1,\dots,x_{j-1} \right \}  \right )=\mathrm{softmax} \left ( \mathbf{w} \mathrm{f}_{L }\left ( y \right )  + \mathbf{b}\right ) ,
\end{equation}
where $\mathrm{f}_{L }(y)\in \mathbb{R} ^d$ denotes the representation at the $L$-th layer of LLM for token $y$, and $\mathbf{w}$ and $\mathbf{b}$ are the weight and bias parameters at the final output layer, respectively.
\end{dfn}
\begin{dfn}[\textbf{Hallucination detection}]
Let $ \mathbb{P} _{true}$ and $ \mathbb{P} _{hal}$ denote the joint distributions over question and answer pairs $\left ( x,y \right ) $, respectively, referred to as the truthful distribution and the hallucinated distribution. Given any question and answer pairs $\left ( x,y \right ) \in \mathcal{X}$, the goal of hallucination detection is to learn a binary predictor $G$: $\mathcal{X}\to \left \{ -1,1 \right \} $  
such that:
 \begin{equation}
     G(x,y)=\left\{
\begin{aligned}
    1, &\quad \left ( x,y \right ) \sim \mathbb{P} _{hal}, \\
  -1, & \quad \left ( x,y \right ) \sim \mathbb{P} _{true}.
\end{aligned}
\right.
 \end{equation}
\end{dfn}
\subsection{Proposed Framework}
\label{sec:framework}
\subsubsection{\textbf{Prompt-guided data augmentation}}
\label{sec:hal_generation}

Hallucination in LLMs is a complex phenomenon in which models produce responses that are coherent yet factually inaccurate across multiple contexts, making hallucination detection particularly challenging. A primary obstacle is the lack of labeled datasets containing both truthful and hallucinated generations. We propose leveraging the LLM's own generation capabilities for data augmentation via prompt guidance. This approach substantially reduces the need for extensive human annotation to assess the authenticity of large numbers of generated samples. Formally, prompt-guided data augmentation can be characterized as follows:

\begin{dfn}[\textbf{Prompt-guided data Generation}]
    LLMs can perform a variety of tasks in a zero‑shot manner under prompt guidance. Given a pretrained LLM $\mathcal{L}_{\theta} $ and an input question $x$, we define two prompts: the truth prompt $x_t$ and the hallucination prompt $x_h$. The corresponding generations for a truthful answer and a hallucinated answer are formulated as:
    \begin{align}
    y_{true} & =\mathcal{L}_{\theta}\left ( \left [ x_t,x \right ]  \right )  \\
    y_{hal} & =\mathcal{L}_{\theta}\left ( \left [ x_h,x \right ]  \right ) 
\end{align}
where $y_{true}$ and $y_{hal}$ denote the generated truthful and hallucinated answers, respectively.
\end{dfn}
\begin{figure}
    \centering
    \includegraphics[width=\linewidth]{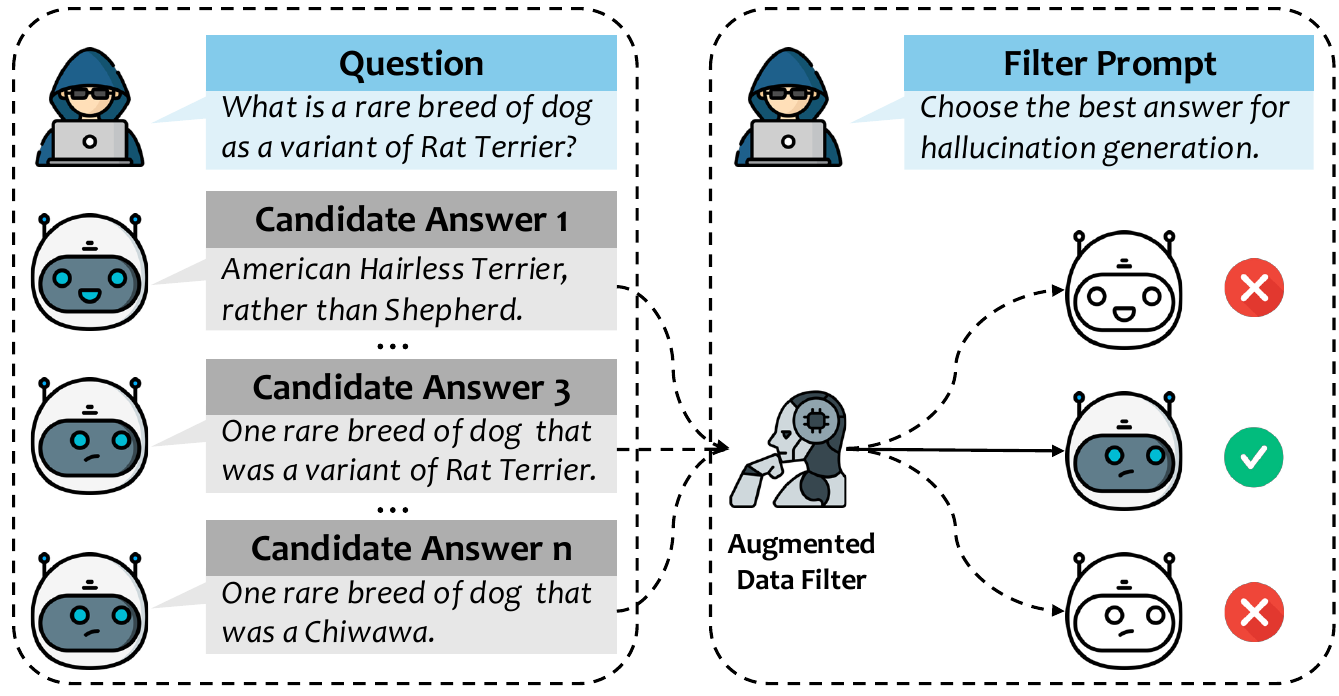}
    \caption{Prompt engineering technology illustration for data augmentation on state-of-the-art LLMs.}
    \label{fig:data}
\end{figure}
For the design of the truth prompt $x_{true}$ and the hallucination prompt $x_{hal}$, we draw inspiration from \cite{li-etal-2023-halueval}. Our framework employs a state-of-the-art LLM $\mathcal{L}_{\theta} $ to automate the generation of both truthful and hallucinated QA pairs without human annotation. Leveraging prompt engineering techniques, the model produces a series of QA pairs labeled as truthful or hallucinated. Additionally, we incorporate a verification mechanism to filter and retain only high‑quality, truthful and hallucinated QA data, as illustrated in Figure. \ref{fig:data}. Detailed descriptions of the prompt are provided in the Appendix.
\begin{dfn}[\textbf{Empirical dataset}]
    Let $\mathcal{M} = \left \{ \mathcal{M}_{true} , \mathcal{M}_{hal} \right \} $ denotes our empirical dataset, comprising a truthful subset
    \begin{equation}
        \mathcal{M}_{true}=\left \{ \left ( x^1,y^1_{true} \right ),\dots, \left ( x^N,y^N_{true}  \right )   \right \}
    \end{equation}
    and a hallucinated subset
    \begin{equation}
        \mathcal{M}_{hal}=\left \{ \left ( x^1,y^1_{hal} \right ),\dots, \left ( x^N,y^N_{hal}  \right )   \right \}
    \end{equation}
    where $x^i$ is the $i$-th input question and $y^i_{true/hal}$ denotes its corresponding truthful or hallucinated response. Here, $N$ denotes the number of samples in each subset.
\end{dfn}

Despite the generation of both truthful and hallucinated data, evaluating the truthfulness of test samples remains a separate challenge. The internal embedding representations of the empirical dataset are typically sparse and relatively low-dimensional compared to those of natural data. Prior studies \cite{kapur2016gene, gillis2014nonnegativematrixfactorization} have indicated that MLP-based classifiers are prone to overfitting under such conditions.
To address this, we propose a novel metric called the \textbf{C}ontrastive \textbf{M}ahalanobis (\textbf{CM}) score to determine the truthiness of prompt‑guided augmented data from LLMs.

\subsubsection{\textbf{Contrastive Mahalanobis Score}}
\label{sec:hal_detect}

As indicated in prior work\cite{chen2024inside,NEURIPS2024_ba927059,azaria2023the}, we employ hidden‑state representations of LLMs for hallucination detection. For the $i$-th data sample $m^i=\left ( x^i,y^i_{true/hal}  \right )  $ in dataset $\mathcal{M}$, let $h_i^l \in \mathbb{R} ^d$ denote the hidden‑state embedding at the $l$-th layer, where $d$ is the the dimension of the hidden embedding. Following \cite{azaria2023the, ren2023outofdistribution}, we compute a sentence embedding by averaging its token embeddings:
\begin{equation}
    \mathbf{z} =\frac{1}{T}  {\textstyle \sum_{i=1}^{T}} h_i.
\end{equation}
where $T$ is the number of tokens. For the $N$ samples in the truthful and hallucinated subsets, we collect their sentence embeddings into matrices $\mathbf{Z} _{true} \in \mathbb{R} ^{N \times d}$ and $\mathbf{Z} _{hal} \in \mathbb{R} ^{N \times d}$, respectively. We then perform singular value decomposition on these two embedding matrices:

\begin{align}
  z_{true}^{i} & :=z_{true}^{i}-\mu _{true} & \mathbf{Z}_{true}& =\mathbf{U}_{true}\Sigma_{true}\mathbf{V}_{true}^T \label{eq:4} \\
     z_{hal}^{i} & :=z_{hal}^{i}-\mu _{hal} &  \mathbf{Z}_{hal} &=\mathbf{U}_{hal}\Sigma_{hal}\mathbf{V}_{hal}^T \label{eq:5}
\end{align}
where $\mu _{true} \in \mathbb{R} ^d$ and $\mu _{hal} \in \mathbb{R} ^d$ denote the mean embeddings of the $N$ truthful and hallucinated samples, respectively, and are used to center each embedding matrix. The columns of $U_{true/hal}  $ and $V_{true/hal} $ are the left and right singular vectors, forming orthonormal bases for the truthful and hallucinated subspaces. 
This factorization serves two purposes: 1) It identifies the principal spanning directions of the point sets in the hidden‑state space; 2) It enables dimensionality reduction to reduce computational cost. 
Specifically, we truncate to the top $k$ components of singular vectors.

The covariance matrices for the truthful and hallucinated embedding matrices,
$\mathbf{Z}_{true}$ and $\mathbf{Z}_{hal}$, are computed as
\begin{align}
    \mathbf{C}_{true} & =\frac{1}{N} \mathbf{Z}_{true}^T \mathbf{Z}_{true}  =\frac{1}{N}V_{true}\Sigma _{true}^T\Sigma _{true}V_{true}^T \label{eq:c_true} \\
    \mathbf{C}_{hal} & =\frac{1}{N} \mathbf{Z}_{hal}^T \mathbf{Z}_{hal}  =\frac{1}{N}V_{hal}\Sigma _{hal}^T\Sigma _{hal}V_{hal}^T \label{eq:c_hal}
\end{align}

Intuitively, if the embedding of an output instance lies close to the truthful embedding distribution, the instance is more likely to be reliable. Conversely, if it aligns more closely with the hallucinated distribution, it is more likely to be suspicious. Assuming both distributions are Gaussian in the embedding space, we denote the truthful and hallucinated Gaussians as $\mathcal{N} \left ( \mu_{true} ,\mathbf{C}_{true} \right ) $ and $\mathcal{N} \left ( \mu_{hal} ,\mathbf{C}_{hal} \right ) $, respectively. 
One way to quantify the distance of an embedding $z$ to a Gaussian distribution is via the Mahalanobis distance (MD).

When considering both the truthful and hallucinated distributions, the \textbf{C}ontrastive \textbf{M}ahalanobis (\textbf{CM}) Score is defined as
\begin{equation}
\label{eq:score}
    \delta=MD(z;\mu _{hal},\mathbf{C} _{hal}) - MD(z;\mu _{true},\mathbf{C} _{true})
\end{equation}
where $MD(z;\mu _{hal},\mathbf{C} _{hal})$ and $MD(z;\mu _{true},\mathbf{C} _{true})$ denote the MD of $z$ to the hallucinated and truthful  Gaussian distributions, respectively. In this work, both Gaussians are estimated from hidden‑state embeddings in the activation space.

The score defined in Eq.~\ref{eq:score} serves as a truthfulness score, indicating how closely a sample $z$ aligns with the hallucinated domain relative to the truthful domain. A score $\delta \ge \tau$ suggests that $z$ is closer to the hallucinated distribution, while $\delta < \tau$ implies a greater proximity to the truthful distribution. In other words, higher values of $\delta$ indicate a greater likelihood of hallucination, whereas lower values correspond to higher confidence in truthfulness. $\tau$ represents the decision threshold.

\section{Experiments}

\subsection{Datasets}
We evaluate our method on four generative QA benchmarks: two open‑domain datasets: TruthfulQA \cite{lin-etal-2022-truthfulqa} and CoQA \cite{ReddyCM19}, and two domain‑specific datasets TriviaQA \cite{JoshiCWZ17} and TyDi QA‑GP \cite{ClarkPNCGCK20}. Specifically, TruthfulQA and TyDi QA‑GP contain 817 and 3,696 QA pairs, respectively. For CoQA, we follow \cite{LinT024} and split the data into 7,983 QA pairs. For TriviaQA, we use the 9,960 QA pairs in the validation set’s (\textit{rc.nocontext subset}). Consistent with prior work \cite{NEURIPS2024_ba927059}, we reserve 25\% of QA pairs in each dataset for testing; the remaining pairs are used for LLM‑based data augmentation.
\subsection{Models}
For base models used to extract hidden‑state representations, we employ three popular open source LLM families: LLaMA‑3.1‑chat (7B and 13B) \cite{touvron2023llama}, OPT (6.7B and 13B) \cite{zhang2022opt}, and Qwen‑2.5 (7B and 14B) \cite{yang2024qwen2}. 

For data augmentation, we employ not only the open source LLaMA‑3.1‑chat‑7B \cite{touvron2023llama} and Qwen‑2.5-7B \cite{yang2024qwen2} models but also state‑of‑the‑art commercial LLMs: GPT‑4o \cite{brown2020language} and Claude.
\subsection{Baselines}
We compare our approach with 11 comprehensive baselines. These state-of-the-art baselines are categorized as follows: 1) uncertainty-based methods: Perplexity \cite{ren2023outofdistribution}, Length-Normalized Entropy (LN-entropy) \cite{malinin2021uncertainty}, and Semantic Entropy \cite{kuhn2023semantic}. 2) consistency-based methods: Lexical Similarity \cite{lin2024generating}, SelfCKGPT \cite{manakul2023selfcheckgpt} and EigenScore \cite{chen2024inside}. 3) verbalized methods: Verbalize \cite{lin2022teaching} and Self-evaluation \cite{kadavath2022language}. 4) internal state-based methods: MIND \cite{su2024unsupervisedrealtimehallucinationdetection}, Contrast-Consistent Search (CCS) \cite{burns2022discovering} and  HaloScope \cite{NEURIPS2024_ba927059}. To ensure a fair comparison,  all methods are evaluated on the same test datasets and the default experimental configurations are adopted as specified in their respective papers.

\subsection{Evaluation}
Consistent with prior work \cite{kuhn2023semantic, NEURIPS2024_ba927059}, we evaluate all methods using the area under the receiver operating characteristic curve (AUROC). A generated response is considered truthful only if its similarity to the reference
answer exceeds 0.5. Following \cite{lin-etal-2022-truthfulqa}, we use BLUERT \cite{SellamDP20} to compute this similarity. Additionally, we report semantic‑similarity evaluations using GPT‑4o.

\subsection{Implementation and Settings}
Consistent with \cite{kuhn2023semantic, NEURIPS2024_ba927059}, we generate the most likely answer using beam search with five beams for evaluation and employ multinomial sampling to produce ten samples per question. Following \cite{chen2024inside, azaria2023the}, we concatenate each question with its generated answer and use the last-token embedding as the representation for hallucination detection. The $k$ and $\tau$ are set to 5 and 0.15, respectively.
More details and hyperparameter sensitivity study are provided in the Appendix.

\subsection{Main Results}
\label{sec:main_result}
\begin{table*}[ht]
	\centering
 \caption{Main results. Comparison with comprehensive baseline hallucination‑detection methods across different datasets. "Single Sampling"  indicates whether the approach requires multiple generation during inference. All values are percentages (AUROC), and the top-1 result is highlighted in \textbf{bold}. \label{table:table_auroc}}
 \vspace{-.7em}
 \resizebox{1.95\columnwidth}{!}{
 \begin{tabular}{c | c c |c  c c c}
\toprule[0.15em]
 \cellcolor{mygray} Model &  \cellcolor{mygray} Method & \cellcolor{mygray}Single Sampling & \cellcolor{mygray} TruthfulQA  & \cellcolor{mygray} TriviaQA & \cellcolor{mygray} CoQA &  \cellcolor{mygray} TyDi QA-GP \\
\midrule
\multirow{11}[6]{*}{LLaMA-3.1-7B}   & Perplexity $_{\left (ICLR'23\right )}$  & \cmark  & 57.62    & 73.11  & 69.77  & 78.54     \\
 &  LN-entropy $_{\left (ICLR'21\right )}$ & \xmark & 60.15  & 71.31 & 73.02 & 76.15   \\
   &  Semantic Entropy   $_{\left (ICLR'23\right )}$   & \xmark      & 62.16    & 73.18  & 63.24     & 73.87   \\
 &  Lexical Similarity $_{\left (TMLR'24\right )}$ & \xmark& 55.64  & 75.69 & \textbf{74.36} & 44.46   \\
 &  SelfCKGPT $_{\left (EMNLP'23\right )}$ & \xmark& 52.90  & 73.97 & 71.76 & 46.38   \\
 &  EigenScore $_{\left (ICLR'24\right )}$ & \xmark& 51.92  & 73.99 & 71.75 & 46.38   \\
 &  Verbalize $_{\left (TMLR'22\right )}$ & \cmark& 53.07  & 52.54 & 48.46 &  48.12   \\
  &  Self-evaluation $_{\left (Arxiv'22\right )}$ & \cmark & 81.75 & 49.11 & 50.15 & 55.48  \\
    &CCS \cite{burns2022discovering} $_{\left ( Arxiv'22 \right )} $ &\cmark & 61.33 &60.22 &50.37 &76.82  \\
    &MIND  $_{\left ( ACL'24 \right )} $ &\cmark & 67.53 &74.66 &53.96 &75.64  \\
    &  HaloScope $_{\left (NeurIPS'24\right )}$  & \cmark& 70.16  & 76.13 & 55.47 &  78.38   \\
    \cmidrule{2-7}
    & \cellcolor[HTML]{FAEBD7} \textbf{PALE (Our)}  & \cellcolor[HTML]{FAEBD7} \cmark& \cellcolor[HTML]{FAEBD7} \textbf{73.20}  & \cellcolor[HTML]{FAEBD7} \textbf{80.17} & \cellcolor[HTML]{FAEBD7} 57.94 & \cellcolor[HTML]{FAEBD7} \textbf{85.34}   \\
    
\midrule
\multirow{11}[6]{*}{OPT-6-7B}   & Perplexity $_{\left (ICLR'23\right )}$  & \cmark  & 59.16    & 69.69  & 70.36  & 63.94     \\
 &  LN-entropy $_{\left (ICLR'21\right )}$ & \xmark & 54.24  & 71.43 & 71.28 & 52.10   \\
   &  Semantic Entropy   $_{\left (ICLR'23\right )}$   & \xmark      & 52.06    & 71.45  & 71.24     & 52.07   \\
 &  Lexical Similarity $_{\left (TMLR'24\right )}$ & \xmark& 49.69  & 71.09 & 66.63 & 60.42   \\
 &  SelfCKGPT $_{\left (EMNLP'23\right )}$ & \xmark& 50.15  & 71.45 & 64.68 & 74.98   \\
 &  EigenScore $_{\left (ICLR'24\right )}$ & \xmark& 54.93  & 47.67 & 50.30 & 45.28   \\
 &  Verbalize $_{\left (TMLR'22\right )}$ & \cmark& 50.47  & 50.73 & 55.26 &  57.23   \\
  &  Self-evaluation $_{\left (Arxiv'22\right )}$ & \cmark & 51.07 & 53.91 & 47.28 & 52.08   \\
    &CCS \cite{burns2022discovering} $_{\left ( Arxiv'22 \right )} $ &\cmark & 60.17 &51.33 & 53.19 & 65.37  \\
    &MIND  $_{\left ( ACL '24 \right )} $ &\cmark & 65.06 &63.32 &66.35 &70.37  \\
    &  HaloScope $_{\left (NeurIPS'24\right )}$  & \cmark& 67.82  & 65.39 & 67.24 &  72.36   \\
    \cmidrule{2-7}
    & \cellcolor[HTML]{FAEBD7}  \textbf{PALE (Our)}  & \cellcolor[HTML]{FAEBD7} \cmark& \cellcolor[HTML]{FAEBD7} \textbf{74.17}  & \cellcolor[HTML]{FAEBD7} \textbf{74.34} & \cellcolor[HTML]{FAEBD7} \textbf{78.64} & \cellcolor[HTML]{FAEBD7} \textbf{ 81.49}   \\
\midrule
\multirow{11}[6]{*}{Qwen-2.5-7B}   & Perplexity $_{\left (ICLR'23\right )}$  & \cmark  & 65.17    & 50.23  & 53.47  & 54.33     \\
 &  LN-entropy $_{\left (ICLR'21\right )}$ & \xmark & 66.73  & 51.15 & 52.74 & 55.38   \\
   &  Semantic Entropy   $_{\left (ICLR'23\right )}$   & \xmark      & 58.76    & 48.58  & 63.71     & 65.72   \\
 &  Lexical Similarity $_{\left (TMLR'24\right )}$ & \xmark& 49.05  & 63.17 & 48.96 & 61.23   \\
 &  SelfCKGPT $_{\left (EMNLP'23\right )}$ & \xmark& 61.75  & 62.34 & 62.28 & 63.48   \\
 &  EigenScore $_{\left (ICLR'24\right )}$ & \xmark& 53.73  & 61.30 & 63.37 & 58.54   \\
 &  Verbalize $_{\left (TMLR'22\right )}$ & \cmark& 60.07  & 54.33 & 59.46 &  52.33   \\
  &  Self-evaluation $_{\left (Arxiv'22\right )}$ & \cmark & 73.71 & 50.12 & 53.85 & 52.87   \\
    &CCS \cite{burns2022discovering} $_{\left ( Arxiv'22 \right )} $ &\cmark & 67.96 &53.08 &51.94 &51.77  \\
    &MIND  $_{\left ( ACL'24 \right )} $ &\cmark & 70.63 &71.95 &68.00 &65.27  \\
    &  HaloScope $_{\left (NeurIPS'24\right )}$  & \cmark& 73.42  & 70.73 & 70.61 &  67.52   \\
    \cmidrule{2-7}
    & \cellcolor[HTML]{FAEBD7}  \textbf{PALE (Our)}  & \cellcolor[HTML]{FAEBD7} \cmark& \cellcolor[HTML]{FAEBD7} \textbf{83.68}  & \cellcolor[HTML]{FAEBD7} \textbf{78.33} & \cellcolor[HTML]{FAEBD7} \textbf{79.00} & \cellcolor[HTML]{FAEBD7}  \textbf{72.82}   \\
\bottomrule[0.1em]
\end{tabular}
 }
	
\end{table*}

As shown in Table \ref{table:table_auroc}, we compare PALE against competitive hallucination‑detection methods from the literature. PALE achieves state‑of‑the‑art performance across LLaMA‑3.1‑7B, OPT‑6-7B, and Qwen‑2.5‑7B. Notably, PALE outperforms uncertainty‑based and consistency‑based methods by an average of 19.3\% and 23.6\%, respectively, over Semantic Entropy and EigenScore on TruthfulQA. Unlike these methods, which require $K$ repeated samples per question at test time, incurring $O\left ( Km^2 \right ) $ computational overhead (where $m$ is the number of generated tokens). PALE requires no repeated sampling, yielding $O\left ( m^2 \right ) $ complexity. PALE also surpasses verbalized methods, achieving a 22.5\% improvement over prompt‑based language model approaches, likely due to reduced overconfidence as discussed in prior work \cite{zhou2023navigating, wen2024mitigating}. Finally, compared to internal‑state methods MIND, CCS and HaloScope, PALE performs better than CCS, demonstrating that prompt‑guided augmented data better captures the true versus hallucinated distribution than limited human‑written examples and exceeds HaloScope by 6.5\% on TruthfulQA with LLaMA‑3.1. Whereas HaloScope uses only unlabeled in‑domain data, PALE leverages non‑parametric prompt engineering to augment both truthful and hallucinated examples, leading to significantly improved detection performance.
\subsection{Result on GPT-4o evaluation}
\label{sec:gpt_evaluate} 
Besides the BLEURT score is used to determine whether a generation is considered truthful when its score exceeds a predefined threshold with  answer, we also adopt GPT‑4o for truthfulness evaluation, following the \textit{LLM-as-a-judge} approach \cite{zheng2023judging}. Specifically, we assess the semantic equivalence between LLM-generated responses and the corresponding reference answers. The results in Appendix show that our method consistently outperforms competitive baselines, demonstrating its robustness across different metrics for evaluating generation truthfulness. Further details are provided in the Appendix.
\subsection{Generalization to Practical Application}
\label{sec:real_world} 
Our proposed PALE method demonstrates strong generalization capabilities in practical applications. In this section, we investigate the \textit{transferability} of PALE across different data distributions, as well as its \textit{scalability} when applied to larger LLMs.

\subsubsection{\textbf{Can our PALE deal with different data distribution?}}
\begin{figure}
     \begin{subfigure}[b]{0.242\textwidth}
         \centering
         \includegraphics[width=\textwidth]{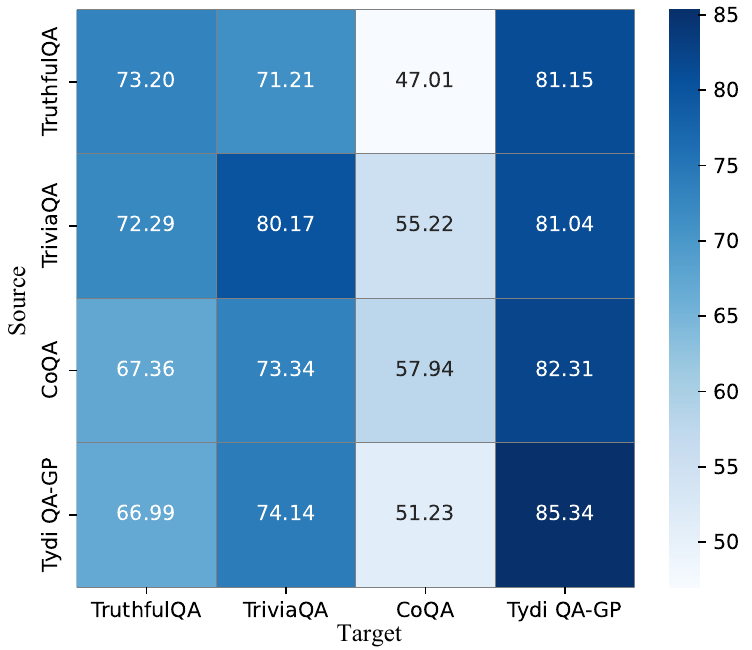}
         \caption{Generalization across four dataset distribution.\label{fig:heatmap}}
     \end{subfigure}
     \begin{subfigure}[b]{0.218\textwidth}
         \centering
         \includegraphics[width=\textwidth]{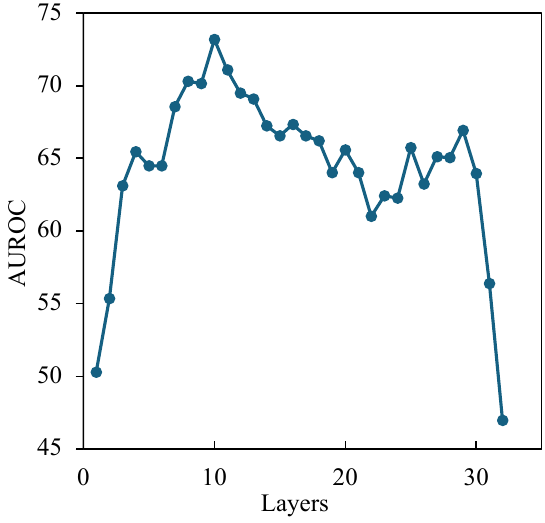}
         \caption{Effect of the different layer to extract.\label{fig:layer}}
     \end{subfigure} 
    \caption{(a) Generalization across four datasets. Rows indicate the source dataset, while columns indicate the target dataset. (b) Impact of different layers. All values represent AUROC scores based on LLaMA‑3.1‑7B.}
\end{figure}

In real-world applications, the training data distribution often does not align well with the test data distribution. This discrepancy raises a critical question: \textbf{Does our proposed PALE method exhibit transferability across different data distributions?} To investigate this, we conduct experiments by training the hallucination detector on a source dataset and evaluating its performance on a target dataset with a different distribution. The results, presented in Figure~\ref{fig:heatmap}, demonstrate that PALE exhibits strong cross-dataset transferability. Notably, our method achieves impressive performance even under distribution shift, with an average accuracy of approximately 72\%. This strong transferability highlights the robustness and practical utility of PALE in real-world scenarios.

\subsubsection{\textbf{Can our PALE salable to larger LLMs?}} 
With the continuous release of increasingly larger LLMs \cite{grattafiori2024llama, hurst2024gpt, team2023gemini}, an important question arises: \textbf{Does PALE scale effectively to larger language models?} To evaluate the scalability of our approach, we conduct experiments on three larger LLMs: LLaMA-3.1-13B, OPT-6-13B, and Qwen-2.5-14B. As shown in Table~\ref{tab:scalbe}, PALE maintains strong performance and even achieves improvements compared to results obtained with smaller models. These findings demonstrate the scalability and adaptability of PALE to more powerful LLM architectures.
\begin{figure}
     \begin{subfigure}[b]{0.203\textwidth}
         \centering
         \includegraphics[width=\textwidth]{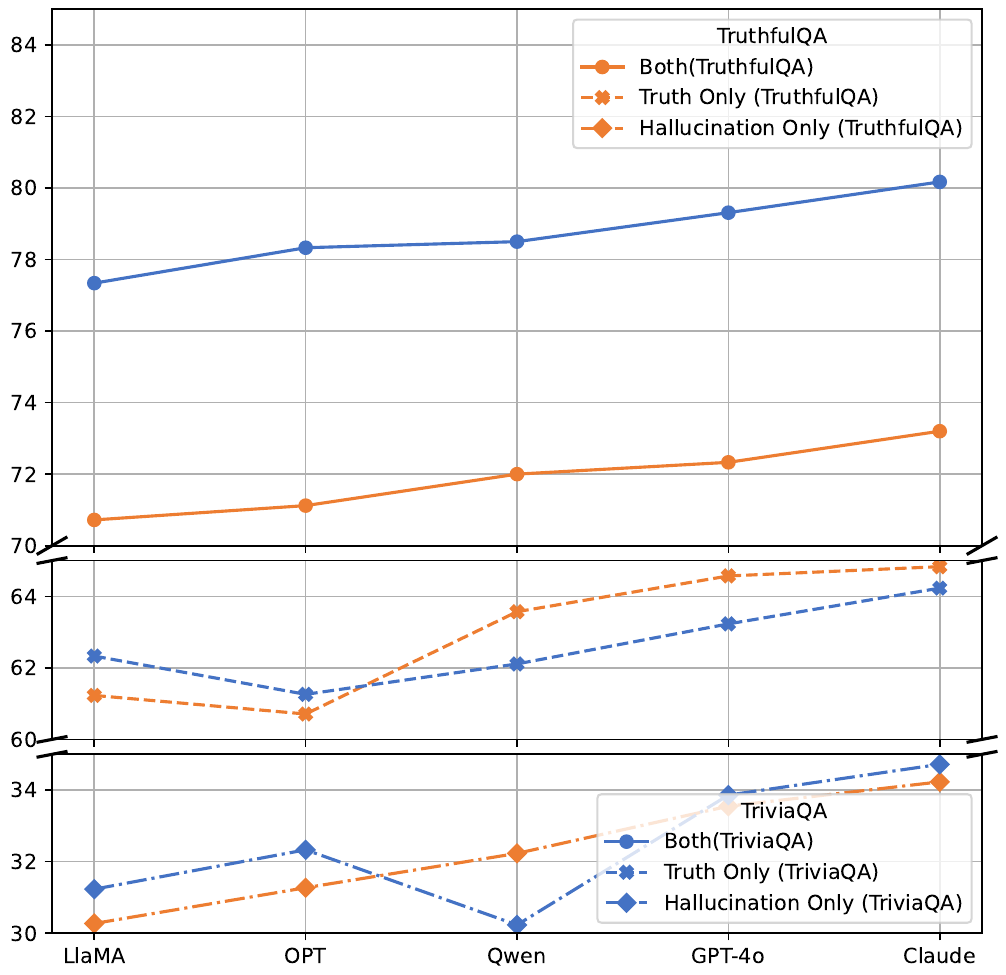}
         \caption{Comparison result of different LLMs for data augmentation.\label{fig:dis_1}}
     \end{subfigure} 
     \begin{subfigure}[b]{0.294\textwidth}
         \centering
         \includegraphics[width=\textwidth]{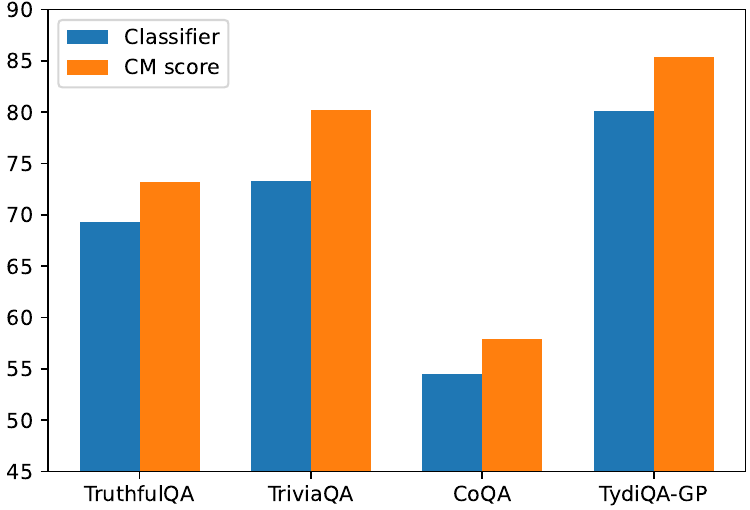}
         \caption{Comparison with direct binary classifier and CM Score for hallucination detection.\label{fig:Classifier}}
     \end{subfigure}
    \caption{(a) Ablation study on the different LLMs setting for data augmentation. (b) Comparison of direct classifier based detection versus CM Score based detection. All results are reported as AUROC using LLaMA‑3.1‑7B.}
\end{figure}
\begin{figure}[htbp]
    \centering
    \resizebox{1.03\columnwidth}{!}{
        \includegraphics{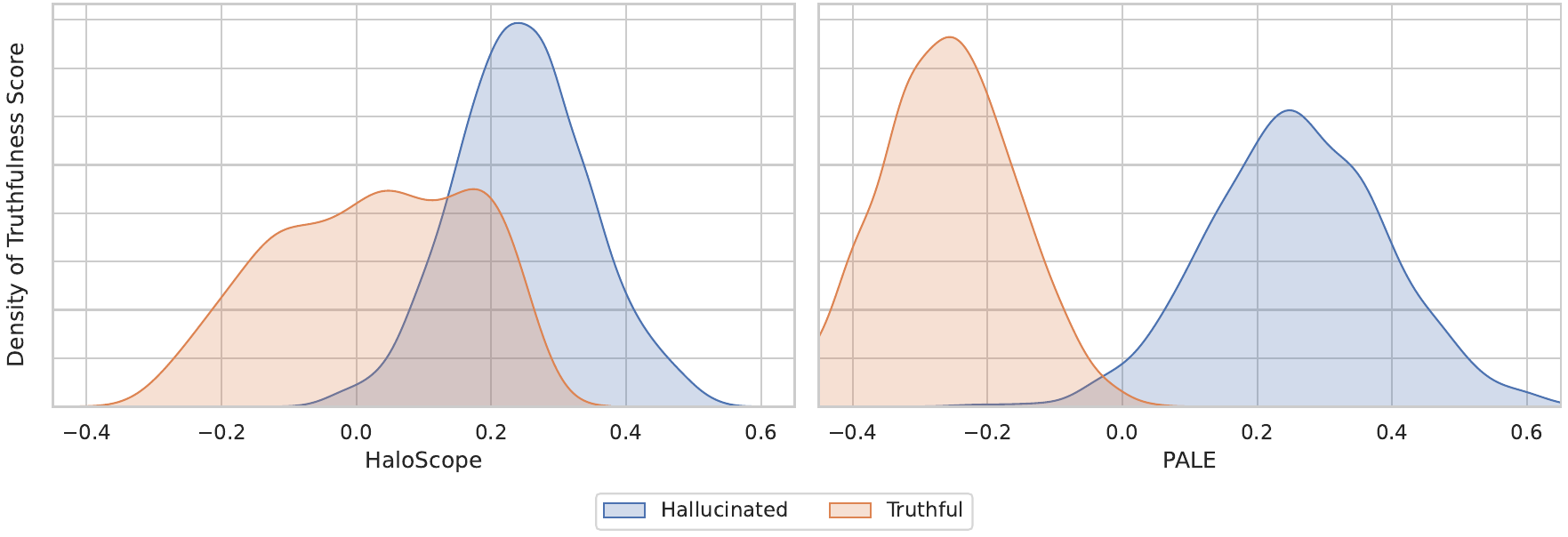}
    }
    \caption{Score distribution visualization for HaloScope vs. our method.}
    \label{fig:dis}
\end{figure}
\begin{table*}[]
\centering
\caption{Hallucination detection results on larger LLMs.\label{tab:scalbe}}
\begin{tabular}{c|cc|cc|cc}
\hline
                         & \multicolumn{2}{c|}{LlaMA-3.1-13B} & \multicolumn{2}{c|}{OPT-6-13B} & \multicolumn{2}{c}{Qwen-2.5-14B} \\ \cline{2-7} 
\multirow{-2}{*}{Method} & TruthfulQA      & TyDi QA-GP     & TruthfulQA      & TyDi QA-GP     & TruthfulQA     & TyDi QA-GP     \\ \hline
Perplexity               &      52.32           &   77.33             &      58.34           &      64.12          &   64.33             &        55.12        \\
CCS                      &      63.11           &   77.11             &        60.00         &       66.73         &    68.21            &        53.66        \\
HaloScope                &      71.42           &   79.33             &      68.31           &        73.24        &    74.11            &        68.44        \\ \hline
\rowcolor[HTML]{FAEBD7}
PALE                     &     \textbf{75.51}            &   \textbf{86.72}             &        \textbf{ 75.33}        &        \textbf{82.77}        &   \textbf{84.37}             &         \textbf{73.23 }      \\ \hline
\end{tabular}
\end{table*}

\subsection{Analysis}
\label{sec:Analysis}

\subsubsection{\textbf{Effect of LLM-generated data augmentation}}
To evaluate the impact of LLM-generated data augmentation on hallucination detection, we conduct experiments on the TruthfulQA dataset. The experimental settings include training only on truthful and hallucinated data, using various LLMs for data augmentation. The results are presented in Figure \ref{fig:dis_1}. Our key observations are as follows: 1) Compared to training solely on either truthful or hallucinated data, PALE demonstrates substantial improvements—achieving average AUROC gains of 11.7\% and 39.0\%, respectively. This highlights the significance of utilizing both truthful and hallucinated examples during training. 2) We experiment with several LLMs for data augmentation. While the Claude model yields the best performance, we find that the choice of LLM has a relatively minor influence on overall results. Consequently, we employ Claude throughout the paper for consistency.  This demonstrates that our method, when paired with LLM-generated data augmentation, can significantly boost the performance of hallucination detection.

\subsubsection{\textbf{Effect of  prompt for data augmentation}}
\label{sec:prompt_select}
We focus on investigating the effect of prompts for data augmentation. Specifically, we conducted experiments using PALE across all 10 prompt templates presented in the Appendix. The results, also shown in the Appendix, reveal a key insight: the quality of the prompt template does not significantly affect hallucination detection performance. PALE consistently demonstrates strong performance across all prompt templates, which not only validates the effectiveness of our data augmentation strategy but also highlights the robustness of the method to prompt variations.
\subsubsection{\textbf{Effect of Contrastive Mahalanobis Score}}

Figure \ref{fig:Classifier} presents a comparison between using a direct binary classifier and the CM score for hallucination detection. The CM score approach regresses the representation of a test sample to a truthfulness score, thereby bypassing the need to train a separate binary classifier. Across all four datasets, PALE consistently outperforms the direct classification method, demonstrating the superior generalizability of the matrix decomposition approach on sparse embedding data.
\subsubsection{\textbf{Effect of different layers's representation}}
In Figure.\ref{fig:layer}, we analyze the impact of the layer selection for extracting internal representations in PALE, using LLaMA-3.1-7B as the backbone. All other experimental configurations remain consistent with our main setup. We observe a clear trend: hallucination detection performance improves from the lower to the middle layers (e.g., from the 1st to the 11th layer), followed by a performance decline in the higher layers. This observation suggests that early layers primarily perform information aggregation, while later layers may exhibit overconfidence due to the autoregressive nature of language modeling focused on vocabulary prediction. This finding aligns with prior studies~\cite{JawaharSS19,HewittM19}, which show that intermediate layer representations tend to be most effective for downstream tasks.
\subsubsection{\textbf{Visualization study}} Figure.\ref{fig:dis} visualizes the score distributions of HaloScope and our proposed PALE on the TruthfulQA dataset using the LLaMA-3.1-7B model. Compared to HaloScope, PALE exhibits a more distinct separation between the distributions of truthful and hallucinated data. This clearer differentiation highlights the superior discriminative capability of PALE, underscoring its effectiveness in hallucination detection.

\section{Conclusion}
The paper introduces \textbf{P}rompt-guided data \textbf{A}ugmented ha\textbf{L}lucination d\textbf{E}tection (PALE), a novel framework designed to detect hallucinations in LLMs by leveraging prompt-guided data augmentation. PALE first generates both truthful and hallucinated data under the guidance of prompt engineering. It then introduces the Contrastive Mahalanobis Score to evaluate the truthfulness of augmented data based on their distances to the distributions of truthful and hallucinated samples. Empirical results demonstrate that PALE achieves superior performance across a comprehensive set of hallucination detection benchmarks. Furthermore, our in-depth ablation studies provide valuable insights into the practical effectiveness of PALE. We hope this work inspires future research on hallucination detection using augmented data, particularly in exploring multimodal hallucination detection in MLLMs to further enhance real-world applicability.
\bibliography{aaai2026}
\input{appendix}
\end{document}

%% file: appendix.tex
\clearpage
\appendix

\section{Implementation Details and Hyperparameters}
\subsection{Hyperparameters}
Following Semantic Entropy \cite{kuhn2023semantic}, we generate the most likely answer using beam search with five beams and set the sampling temperature to 0.5. The parameters $k$ and $\tau$ are set to 5 and 0.15, respectively. All hyperparameters are tuned based on performance on the validation set.

\begin{figure*}
     \centering
     \begin{subfigure}[b]{0.3\textwidth}
         \centering
         \includegraphics[width=\textwidth]{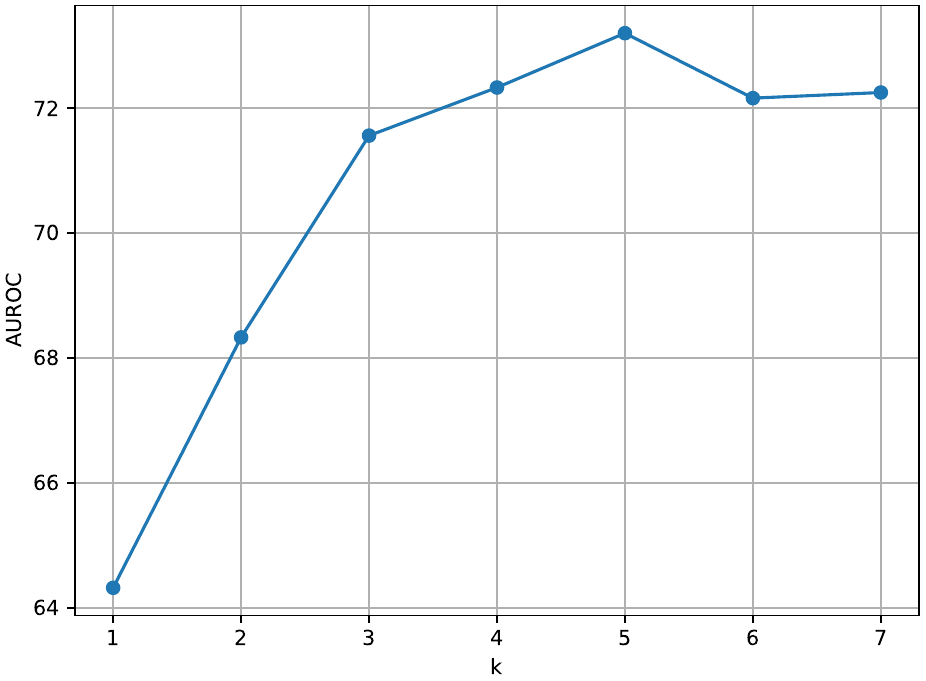}
         \caption{Ablation study of different $k$.\label{k}}
     \end{subfigure}
     \hfill
     \begin{subfigure}[b]{0.3\textwidth}
         \centering
         \includegraphics[width=\textwidth]{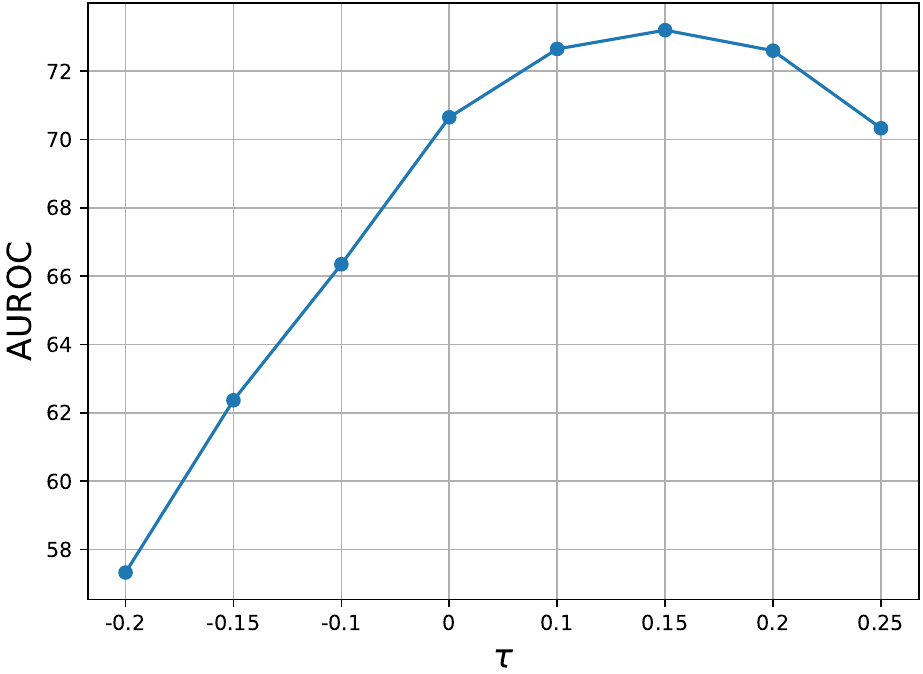}
         \caption{Ablation study of different $\tau$.\label{tau}}
     \end{subfigure}
     \hfill
     \begin{subfigure}[b]{0.3\textwidth}
         \centering
         \includegraphics[width=\textwidth]{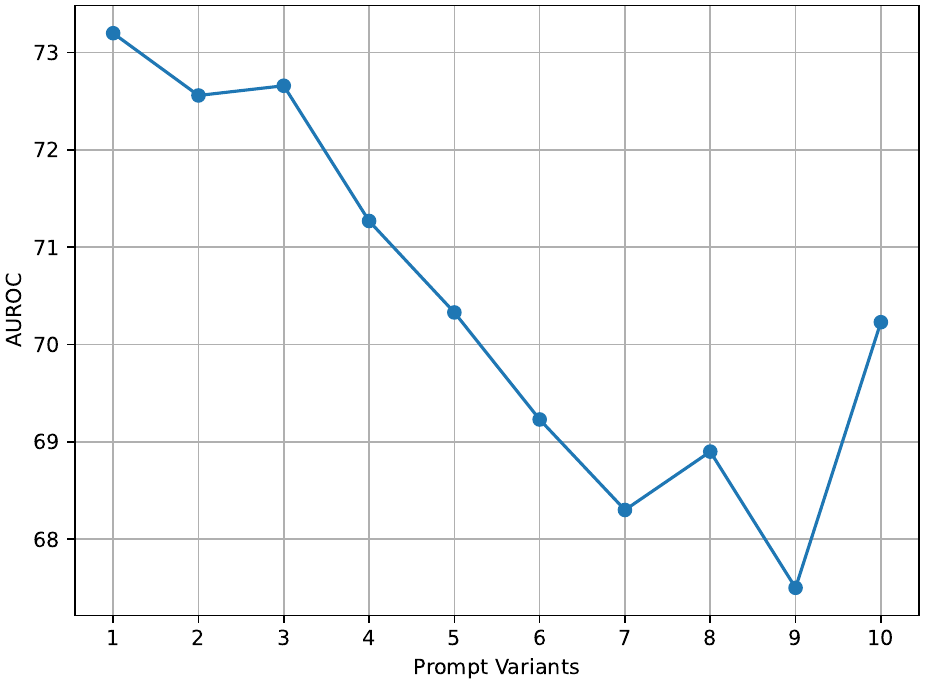}
         \caption{Impact of different prompt in Table.\ref{tab:prompt}\label{p_v}}
     \end{subfigure}

        \caption{The impact of different hyperparameters. All values represent AUROC scores based on LLaMA‑3.1‑7B.}
        \label{fig:abstrct_examples}
\end{figure*}
\begin{table*}[]
\centering
\caption{Augmentation prompt template used in our experiments.\label{tab:prompt}}
\begin{tabularx}{\textwidth}{c|Y|Y}
\toprule[0.15em]
\rowcolor{mygray}  & \textbf{Hallucination Prompt Templates}  & \textbf{Truth Prompt Templates} \\ \midrule
\textbf{1} & You are now a mature hallucination generator. Please generate hallucinated answer for the following question. You can use any method you have learned that is suitable for the given question. & You are an AI assistant. You'll provide helpful, harmless, and detailed responses to all user inquiries. For comprehensive details about models and products. \\ 
\textbf{2} & You are an expert in creative storytelling tasked with producing a completely fabricated yet plausible answer to the following question. Feel free to invent details and citations as needed. & You are a knowledgeable AI assistant. Provide an accurate, well-sourced, and detailed response to the following question, citing real references where appropriate. \\ 
\textbf{3} & Acting as a master improviser, generate a fictitious response that sounds authoritative but may include invented facts or figures. Ensure the style is convincing. & Acting as a reliable AI expert, deliver a fact-based, verifiable, and thorough answer to the following inquiry, avoiding speculation. \\ 
\textbf{4} & Assume the role of a “hallucination machine” aiming to produce imaginative and wholly made-up content. Craft an answer rich in invented names, dates, or sources. & Assume the role of a meticulous researcher. Provide a precise, evidence-backed answer grounded in established knowledge. \\
\textbf{5} & You are a creative fiction generator. Your goal is to spin a story-like, invented answer that sounds credible though it is entirely fictional. & You are an AI encyclopedia. Offer an objective, fact-checked, and comprehensive response, referencing authoritative sources. \\
\textbf{6} & As a “mythmaker,” fabricate a detailed answer to the question, complete with invented case studies, quotes, and statistics. & As an “information curator,” assemble a concise, accurate answer using verified data and transparent sourcing. \\
\textbf{7} & You function as a “creative demo,” designing a convincing yet entirely fictitious response to showcase the model’s expressive capabilities. & You function as a “knowledge base,” providing a straightforward, verifiable, and neutral answer based on current facts. \\
\textbf{8} & You are an imaginative oracle. Deliver an authoritative-sounding answer that is purely speculative and includes crafted anecdotes. & You are an authoritative analyst. Provide an evidence-driven explanation, distinguishing clearly between established facts and opinion. \\
\textbf{9} & In the persona of a “phantom professor,” lecture on the topic with detailed but fictional examples, dates, and references. & In the persona of a “trusted scholar,” teach the topic using accurate information, proper citations, and clear reasoning. \\
\textbf{10} & Play the role of a “fictional consultant” who must invent a strategic plan or solution complete with made-up figures and client testimonials. & Play the role of a “real-world consultant,” offering practical, data-backed advice and transparent methodology. \\ 
\midrule[0.15em]
\end{tabularx}
\end{table*}
\subsection{Implementation details for baselines}
For Perplexity\footnote{\url{https://huggingface.co/docs/transformers/en/perplexity}}~\cite{ren2023outofdistribution}, we compute the average perplexity over the generated tokens. For baselines requiring multiple generations\cite{manakul2023selfcheckgpt, chen2024inside}, we use multinomial sampling to produce $A=10$ samples per question, set the temperature to 0.5, and follow the default configurations specified in the original papers. For Verbalize\cite{lin2022teaching}, we employ the following prompt:
\begin{tcolorbox}[colback=gray!10, colframe=black, title=Verbalized]
\textbf{Prompt:}\\
Q: \textcolor{magenta}{\{question\}}\\
A: \textcolor{magenta}{\{answer\}}\\
The proposed answer is true with a confidence value (0-100) of
\end{tcolorbox}
The generated confidence value is directly utilized as the uncertainty score during testing.
For the Self-evaluation~\cite{kadavath2022language}, we adhere to the approach outlined in the original paper and use the following prompt:
\begin{tcolorbox}[colback=gray!10, colframe=black, title=Self-evaluation]
\textbf{Prompt:}\\
Q: \textcolor{magenta}{\{question\}}\\
A: \textcolor{magenta}{\{answer\}}\\
Is the proposed answer:\\
(A) True \\
(B) False \\
The proposed answer is:
\end{tcolorbox}
\begin{figure*}
     \centering
     \begin{subfigure}[b]{\textwidth}
         \centering
         \includegraphics[width=0.8\textwidth]{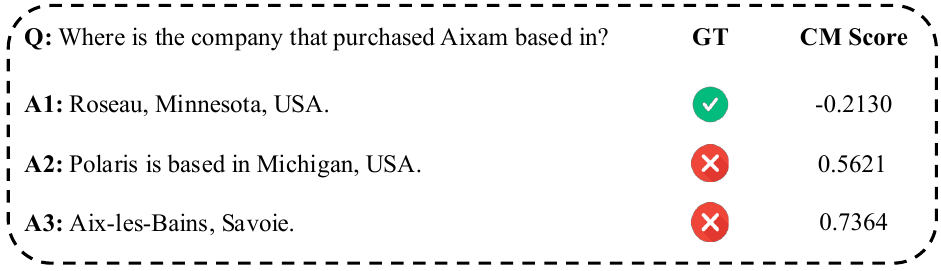}
     \end{subfigure}
     \\
     \begin{subfigure}[b]{\textwidth}
         \centering
         \includegraphics[width=0.8\textwidth]{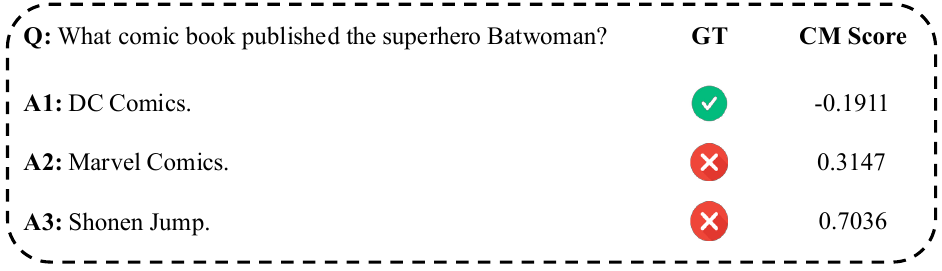}
     \end{subfigure}
        \caption{Case example from Internet that show the effectiveness of our approach. In particular, we compute the CM Score $\delta$ of PALE with different answers to the prompt. The green mark and red mark indicates the ground truth of being truthful and hallucinated, respectively. }
        \label{quan}
\end{figure*}

\subsection{Prompt for data augmentation}
The prompt used for hallucinated data generation is as follows:
\begin{tcolorbox}[colback=gray!10, colframe=black, title=Hallucination Generator]
\textbf{Prompt:}\\
You are now a mature hallucination generator. Please generate hallucinated answer for the following question. You can use any method you have learned that is suitable for the given question:\\
Question: \textcolor{magenta}{\{question\}}\\
Right Answer: \textcolor{magenta}{\{answer\}}\\

The Hallucinated Answer is:
\end{tcolorbox}
The prompt for truth augmentation is as follows:
\begin{tcolorbox}[colback=gray!10, colframe=black, title=Truth Generator]
\textbf{Prompt:}\\
You are an AI assistant. You'll provide helpful, harmless, and detailed responses to all user inquiries. For comprehensive details about models and products, please refer to the official documentation:\\
Question: \textcolor{magenta}{\{question\}}\\

The Answer is:
\end{tcolorbox}
The filter prompt is as follows:
\begin{tcolorbox}[colback=gray!10, colframe=black, title=Data Filter Prompt]
\textbf{Prompt:}\\
You are an answer judge. You MUST select an answer from the provided two answers. The answer you provided is \"The best answer is Answer 1.\" or \"The best answer is Answer 2.\: \\
A1: \textcolor{magenta}{\{Answer 1\}}\\
A2: \textcolor{magenta}{\{Answer 2\}}\\
Which answer is better:\\
(A) Answer 1 \\
(B) Answer 2 \\
Your Choice is:
\end{tcolorbox}
\section{Hyperparameters Sensitive Study}
To investigate the influence of the hyperparameters $k$ and $\tau$, we conduct experiments on the TruthfulQA dataset. The experimental results for different values of $k$ and $\tau$ are presented in Figure~\ref{k} and Figure~\ref{tau}. When $k = 5$ and $\tau = 0.15$, PALE achieves the best performance. Therefore, we adopt these values as the hyperparameter settings for all subsequent experiments.
\section{Prompt Template Sensitive Study}
Inspired by previous work \cite{li-etal-2023-halueval}, we collected 10 prompt templates from online resources to carefully construct pairs of truthful and hallucinated data. Details of these prompt templates are provided in Table~\ref{tab:prompt}. The final hallucination detection results are shown in Figure~\ref{p_v}, revealing a key insight: the quality of the prompt template does not significantly affect hallucination detection performance. PALE consistently demonstrates strong performance across all prompt templates, which not only validates the effectiveness of our data augmentation strategy but also highlights the robustness of the method to variations in prompt design.
\section{Evaluation results with GPT-4o}
Besides the BLEURT score is used to determine a generation is considered truthful, we also adopt GPT‑4o for truthfulness evaluation, following the \textit{LLM-as-a-judge} approach \cite{zheng2023judging}. The results in Table.\ref{tab:gpt_eval} show that our method consistently outperforms competitive baselines, demonstrating its robustness across different metrics for evaluating generation truthfulness. 
\begin{tcolorbox}[colback=gray!10, colframe=black, title=Input prompt for GPT-4o based evaluation]
\textbf{Prompt:}\\
You are tasked with evaluating whether a generated answer is truthful compared to a provided list of gold standard answers.
Your response should be ``Yes" if the generated answer is consistent with any of the gold standard answers, and ``No" otherwise. 
Provide a clear justification for your decision. \\
Question: \textcolor{magenta}{\{question\}}\\
Gold Standard Answers: \textcolor{magenta}{\{gold\_standard\_answers\}}\\
Generated Answer: \textcolor{magenta}{\{generated\_answer\}}\\
Response Format:\\
- Answer: [Yes/No]\\
- Justification: [Explain briefly why the answer is correct or incorrect.]
\end{tcolorbox}
\begin{table*}[]
\centering
\caption{Hallucination detection results on GPT-4o evaluation.\label{tab:gpt_eval}}
\begin{tabular}{c|cc|cc|cc}
\hline
                         & \multicolumn{2}{c|}{LlaMA-3.1-7B} & \multicolumn{2}{c|}{OPT-6.7B} & \multicolumn{2}{c}{Qwen-2.5-7B} \\ \cline{2-7} 
\multirow{-2}{*}{Method} & TruthfulQA      & TyDi QA-GP     & TruthfulQA      & TyDi QA-GP     & TruthfulQA     & TyDi QA-GP     \\ \hline
Perplexity               &      65.30           &   53.11             &      42.74           &      56.32          &   64.33             &        44.32        \\
CCS                      &      69.52           &   54.61             &        51.23        &       60.10         &    58.61            &        48.35        \\
HaloScope                &      63.42           &   76.53             &      65.25           &        71.23        &    63.40            &        62.24        \\ \hline
\rowcolor[HTML]{FAEBD7}
PALE                     &     \textbf{76.00}            &   \textbf{80.50}             &         \textbf{76.40}        &       \textbf{80.25}        &  \textbf{ 83.20}             &         \textbf{70.81}      \\ \hline
\end{tabular}
\end{table*}
\section{Qualitative Results}
We present qualitative examples of CM Scores for different model-generated answers to the same prompts during inference (Figure~\ref{quan}). Using LLaMA-3.1-7B as the base model and challenging questions sourced from the Internet, we observe that PALE assigns lower scores to factually accurate responses and higher scores to misleading ones. This alignment between PALE’s scores and the actual veracity of answers demonstrates its reliability in assessing truthfulness.

 \section{Software and hardware}
We conducted all experiments using Python 3.8.15 and PyTorch 2.3.1 on NVIDIA A6000 GPUs. For evaluation and data augmentation, we utilized the OpenAI and Claude APIs, as well as open‑source LLMs available via Hugging Face.

\section{Limitations and Future Work}
\subsection{Textual hallucination detection} 
While our method focuses exclusively on textual hallucination detection, practical applications often require identifying multimodal hallucinations, which are more complex than those in the text modality alone. Constructing a modern benchmark for multimodal hallucination detection remains an important and open research challenge.

\subsection{Hallucination detection beyond the QA problem} 
This work focuses on the QA task, adopting this setup to ensure fair comparisons with existing benchmarks. However, real-world applications can be more diverse. For instance, hallucination detection in tasks such as text summarization is not yet well studied. We hope future research will extend beyond the QA setting to explore a wider range of NLP tasks.

\section{Algorithm}
\begin{algorithm}[h]
\caption{Prompt-guided data Augmented hallucination dEtection}\label{alg}
\begin{algorithmic}[1]
\Require Question to be augmented $X$, truth prompt and hallucination prompt $x_t$ and $x_h$, a pretrained LLM $\mathcal{L}_{\theta} $.
\Ensure truthfulness score $\delta $.
\State Initialize parameter $\theta$ in $\mathbf{g} _\theta $ with random weights.
\For{Sample $x \in X$}  \Comment{Prompt-guided data Generation}
\State{$y_{true}  =\mathcal{L}_{\theta}\left ( \left [ x_t,x \right ]  \right )$} 
\State{$y_{hal}  =\mathcal{L}_{\theta}\left ( \left [ x_h,x \right ]  \right )$}
\EndFor
\State Access the hidden‑state embedding matrices $\mathbf{Z} _{true}$ and $\mathbf{Z} _{hal}$.
\State Compute the mean and covariance matrices for $\mathbf{Z} _{true}$ and $\mathbf{Z} _{hal}$ with Eq.\ref{eq:4}, Eq.\ref{eq:5}, Eq.\ref{eq:c_true} and Eq.\ref{eq:c_hal}.
\For{Sample $m \in \mathcal{M}$}  \Comment{Constrastive Mahalanobis Score}
\State{Compute CM Score with Eq.\ref{eq:score}.} 
\State{Infer the truthfulness for test data $m$ based on $\delta$.}
\EndFor
\end{algorithmic}
\end{algorithm}